%% file: main.tex
\documentclass[10pt,twocolumn,letterpaper]{article}
\usepackage[pagenumbers]{cvpr}
\usepackage{multirow}
\usepackage{algorithm}
\usepackage{algorithmic}
\usepackage{placeins}

\definecolor{linkblue}{rgb}{0.21,0.49,0.74}
\usepackage[breaklinks,colorlinks,allcolors=linkblue]{hyperref}

\makeatletter
\newcommand{\fullwidthtitle}[1]{%
  \if@twocolumn
    \twocolumn[{%
      \begin{center}%
        {\Large\bfseries #1\par}%
      \end{center}%
      \vspace{1em}%
    }]%
  \else
    \begin{center}%
      {\Large\bfseries #1\par}%
    \end{center}%
    \vspace{1em}%
  \fi
}
\makeatother

\title{Visual Anchoring in Diffusion: Multimodal Zero-Shot Skeleton Action Recognition}
\author{Zehao Bao\\
The University of Hong Kong\\
Hong Kong SAR, China\\
{\small\ttfamily\detokenize{zehao_bao@connect.hku.hk}}
\and
Shujun Guo\\
Zhejiang University\\
Hangzhou, China\\
{\small\ttfamily\detokenize{shujunguo@zju.edu.cn}}
\and
Bruce X.B. Yu\thanks{Corresponding author.}\\
Zhejiang University\\
Hangzhou, China\\
{\small\ttfamily\detokenize{xinboyu@intl.zju.edu.cn}}
}
\date{}
\hypersetup{
  pdftitle={Visual Anchoring in Diffusion: Multimodal Zero-Shot Skeleton Action Recognition},
  pdfauthor={Zehao Bao, Shujun Guo, Bruce X.B. Yu}
}

\begin{document}
\maketitle

\begin{abstract}
Zero-shot Skeleton Action Recognition (ZSAR) remains ambiguous when unseen actions share similar skeleton joint dynamics but differ in objects or scene context. RGB provides these missing cues, yet existing multimodal methods typically maintain independent skeleton and RGB scoring branches and fuse their outputs. Without using unlabeled test data for adaptation or fusion calibration, a fixed fusion weight cannot capture class-pair-dependent modality reliability, while an adaptive rule lacks target-side feedback for deciding which branch should dominate. We bypass this weight-selection problem via the classify-by-generation paradigm, where each class is scored by how accurately a text-conditioned denoiser predicts the noise added to the skeleton feature. This formulation separates the progressively corrupted skeleton from fixed conditioning, allowing RGB and text to jointly condition a single class-scoring function rather than produce independent scores. We instantiate this idea as Multimodal Triplet Diffusion for Skeleton–Text Matching (TDSM-MM)\footnote{Project page: \url{https://github.com/ZehaoBao/TDSM-MM}}, augmenting a text-conditioned denoising Transformer with a non-diffused RGB condition token that serves as a stable visual anchor during skeleton data reconstruction. Our proposed TDSM-MM has been ablated via extensive experiments and achieved the best inductive accuracy on three of four NTU-60/120 splits and surpasses the transductive state-of-the-art on NTU-120 96/24 (i.e., 71.3\% vs.\ 69.1\%), without test-time adaptation, suggesting that diffusion-based methods can be a promising direction for zero-shot learning.
\end{abstract}

\section{Introduction}

Skeleton-based human action recognition \citep{cheng2020skeleton, shahroudy2016ntu, liu2019ntu} discards background and object appearance to focus on body dynamics, yielding a compact representation that becomes a limitation when actions share similar joint motion and differ only in visual context: a held knife is invisible in skeleton representations, and headache versus salute produce nearly identical poses. Supervised methods tolerate this when every class has labeled examples, but the space of human actions is open-ended, motivating Zero-shot Action Recognition (ZSAR) \citep{gupta2021syntactically, zhou2023zero, do2025bridging}, which recognizes unseen action categories that have no labeled training instances by leveraging their semantic descriptions.

\begin{figure}[!t]
  \centering
  \includegraphics[width=0.75\columnwidth,
   keepaspectratio]{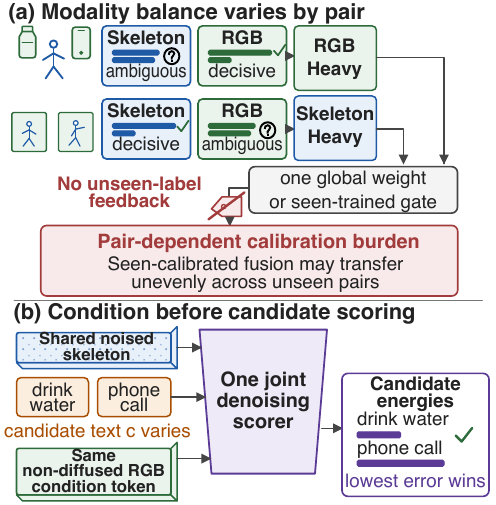}
  \caption{Motivation. (a) Pair-varying modality reliability burdens post-hoc fusion. (b) TDSM-MM jointly scores candidates from skeleton, text, and RGB.}
  \label{fig:motivation}
\end{figure}

ZSAR bridges seen and unseen classes through a shared semantic space, but this space inherits the limitations of the available modalities: when an action is distinguished primarily by appearance rather than motion, text descriptions
cannot fully compensate for the missing visual evidence in skeleton features. RGB therefore provides a natural complementary modality. However, existing RGB-enhanced ZSAR methods have not consistently achieved gains over strong
skeleton-based approaches. For example, Beyond-Skeleton Zero-shot Learning (BSZSL)~\citep{liu2025beyond} does not consistently outperform skeleton-only Triplet Diffusion for Skeleton-Text Matching (TDSM)
\citep{do2025bridging}, indicating that simply introducing RGB features is insufficient and that the modality interaction mechanism remains a critical challenge.

Many multimodal recognition approaches explore modality interaction through feature-level integration, representation alignment, or decision-level fusion
\citep{baltruvsaitis2019multimodal, dong2020survey, bruce2022mmnet}. These strategies generally process modalities as separate branches and combine them after modality-specific representations have been learned. We argue that zero-shot recognition
requires a different perspective: because unseen classes provide no labels for calibrating modality interactions, complementary modalities should participate directly in the recognition process rather than being combined through
post-hoc score fusion. We therefore formulate multimodal ZSAR as a conditioning problem inside a diffusion-based classify-by-generation framework
\citep{li2023your, clark2023text, do2025bridging}: a candidate class is evaluated according to how accurately the denoiser predicts the diffusion noise under the corresponding class and modality conditions. Unlike previous multimodal fusion strategies, this formulation enables the auxiliary modality to guide generation rather than provide an independent prediction score.

RGB provides a natural conditioning signal for two complementary reasons. First, RGB bridges the skeleton--text modality gap: RGB and text are already well aligned through large-scale vision-language pretraining \citep{radford2021learning, cherti2023reproducible}, while RGB and skeleton describe the same action instance from different views, giving the denoiser an intermediate anchor between the two. Second, text conditioning alone is sparse, naming only the class label without the execution style, viewpoint, or object interactions that distinguish visually similar actions; RGB supplies sample-specific evidence unavailable to class-level text, allowing the denoiser to evaluate each candidate under the observed visual context. This raises a specific question, however: why must RGB enter as a non-diffused RGB condition token inside the denoiser, rather than through a more direct alternative such as early fusion into the skeleton representation, independent RGB score fusion, or a conditioning-token interface without a corruption-based objective? We find that each of these alternatives fails to preserve the anchor property motivated above.

We instantiate this as Multimodal Triplet Diffusion for Skeleton–Text Matching (TDSM-MM), extending a text-conditioned denoising transformer with a single non-diffused RGB condition token (Figure~\ref{fig:motivation}): the denoiser attends jointly over the noised skeleton, text tokens, and this RGB condition token, whose stable presence enables it to serve as a visual anchor whose influence grows with the skeleton's corruption, extending to the visual modality the role text plays in text-to-image diffusion \citep{rombach2022high, ho2022classifier}. 
The contributions of this work are threefold:

\begin{itemize}
\item \textbf{Diffusion-based multimodal formulation.}
We introduce the first diffusion-based formulation for multi-modal
zero-shot skeleton action recognition, treating RGB as a condition
within the classify-by-generation framework rather than as an
independent prediction branch.
\item \textbf{RGB conditioning mechanism.}
Based on this formulation, we introduce a non-diffused RGB condition token that remains uncorrupted while the skeleton representation is progressively noised. This design exploits two complementary properties of RGB: bridging the skeleton--text modality gap and providing instance-level visual details unavailable from sparse text descriptions. The non-diffused RGB condition token serves as a stable visual anchor during denoising.
\item \textbf{Empirical validation.}
We instantiate the proposed framework as TDSM-MM, achieving the best inductive accuracy on three of four NTU-60/120 splits and surpassing the transductive state of the art on NTU-120 96/24 without test-time adaptation. Further analysis shows that the improvement mainly comes from object- and posture-dependent unseen classes.
\end{itemize}

\section{Related Work}
\label{sec:related}

\paragraph{Zero-Shot Learning.} Zero-shot learning (ZSL) recognizes target classes without labeled examples for those classes by transferring knowledge from labeled seen classes through auxiliary semantics such as attributes or language~\citep{wang2019zsar}. Its semantic interface has evolved from manually engineered attribute, lexical, and keyword spaces toward learned label and text embeddings, reducing the need for hand-designed representations~\citep{wang2019zsar}. ZSL protocols are commonly divided into inductive and transductive settings according to whether the model exploits target-domain information beyond the candidate-class semantics~\citep{wang2019zsar}. Inductive methods learn from labeled seen-class samples and apply a fixed model to unseen instances. Transductive methods additionally use target-side information at the class or instance level: PGFA uses designated unseen-class prototypes to align text features; DynaPURLS adapts representations using unlabeled target instances accumulated during inference; and Skeleton-Cache builds a nonparametric cache from incoming target samples~\citep{zhou2025zero,zhu2026dynapurls,zhu2026boosting}. We focus on inductive, fixed-model inference under the original TDSM protocol~\citep{do2025bridging}, scoring each sample independently without updating model parameters, prompts, prototypes, or caches from incoming target instances. With this protocol boundary established, the next question is what evidence an action-recognition model observes at inference.

\paragraph{Zero-Shot Action Recognition.} Human action recognition (HAR) has expanded from RGB and gray-scale video to depth, skeleton, inertial, and other sensing modalities~\citep{sun2022human}. Despite this expansion in sensing modalities, conventional systems generally learn a fixed set of actions from labeled examples, whose collection and annotation are costly, particularly for emerging or rare activities~\citep{chen2021deep}. Zero-shot action recognition (ZSAR) instead transfers knowledge from seen to disjoint unseen action categories through shared class semantics~\citep{wang2019zsar,gupta2021syntactically}.

Skeleton-based ZSAR has progressed from POS-aware generative embeddings to mutual-information alignment and dual skeleton--text alignment, body-part and temporal modeling, and context-aware semantics~\citep{gupta2021syntactically,zhou2023zero,kuang2025zero,zhu2024part,chen2025neuron,wang2026skeletoncontext}; despite these advances, it does not observe current-video appearance at inference. Conversely, RGB-centric zero-shot video methods retain appearance and scene cues but do not maintain an explicit skeleton stream for body dynamics~\citep{chen2021elaborative,ni2022expanding,rasheed2023fine}. These complementary limitations motivate multimodal skeleton ZSAR.

Multimodal skeleton methods differ primarily in whether RGB is retained at inference and how it enters the class decision. Training-only learners such as C2VL~\citep{chen2024vision} and MMCL~\citep{liu2024multi} use RGB or vision--language supervision to improve skeleton representations but return to skeleton-only inference. The closest multimodal ZSAR method, BSZSL~\citep{liu2025beyond}, instead retains sample-specific RGB at test time and applies text-guided RGB--skeleton feature enhancement and contrastive alignment before multiplying a visual-semantic class factor by a skeleton prediction factor. It therefore uses RGB through representation enhancement and modality-specific prediction rather than as a condition of a candidate-wise generative scorer. Notably, TDSM reports higher accuracy without RGB on three of the four NTU-60/120 splits~\citep{do2025bridging}, suggesting that the benefit of visual evidence depends not only on its availability but also on how it enters candidate scoring.

\paragraph{Generative Models for Zero-Shot Recognition.} Generative ZSL methods differ in whether generation synthesizes training data or directly supplies a candidate-class score. An established route generates unseen-class feature instances for a downstream classifier~\citep{wang2019zsar,chen2025genzsl}; skeleton-specific VAE variants improve this route through semantic factor disentanglement in SA-DVAE and frequency-semantic enhancement in FS-VAE~\citep{li2024sa,wu2025frequency}. Direct-scoring methods instead rank candidates with a class-conditioned error or energy, as in SCALE~\citep{oraki2026scale}. Diffusion classifiers implement this interface by keeping the same noised observation fixed, substituting each candidate condition in turn, and ranking candidates by their noise-prediction errors~\citep{li2023your,clark2023text}. TDSM applies this rule to skeleton ZSAR by noising a skeleton representation, conditioning the denoiser on each candidate class text, and selecting the text that produces the minimum denoising error~\citep{do2025bridging}; FDSM retains this candidate-wise interface while introducing frequency-aware modeling and curriculum-guided semantics~\citep{zhou2026frequency}. Flora instead uses flow matching and condition-free contrastive regularization to form a token-level velocity classifier~\citep{chen2026learning}. 

These strands expose a scoring-interface gap rather than simply a missing modality combination. Visual-assisted ZSAR shows that appearance from current sample's video can complement skeleton motion, yet existing methods use it either for training-time representation enhancement or for modality-specific prediction. Candidate-wise diffusion offers a different interface by evaluating each class hypothesis through one denoising energy. TDSM-MM therefore places the current RGB representation inside the same denoiser that processes the noised skeleton and candidate text: holding the skeleton and RGB fixed while varying only the candidate text allows visual evidence to modify the relative denoising errors—and therefore the ranking—of candidate classes, without producing an independent RGB ranking or requiring a post-hoc fusion weight.

\section{Method}
\label{sec:method}

\begin{figure*}[t]
\centering
\includegraphics[width=\textwidth]{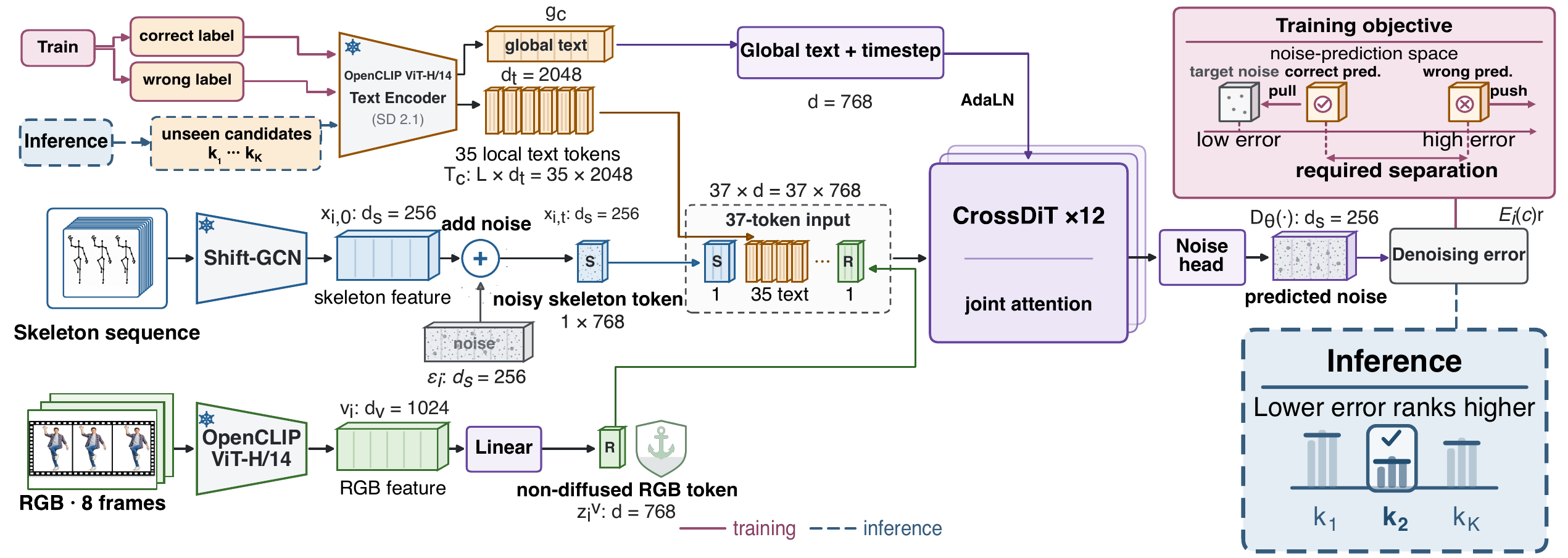}
\caption{Training and inference in TDSM-MM. Only skeleton features undergo forward diffusion; candidate text and current-video RGB jointly condition the CrossDiT denoiser. Training enforces a lower denoising error for the correct text than a wrong text, whereas inference ranks unseen candidates by the same error without an independent RGB score.}
\label{fig:arch}
\end{figure*}
\subsection{Overview and Notation}

\paragraph{Task setting.} Let $\mathcal{Y}_s$ and $\mathcal{Y}_u$ be disjoint sets of seen and unseen action classes. The seen-class training set is $\mathcal{D}_s=\{(\boldsymbol{x}_{i,0},\boldsymbol{v}_i,y_i)\}_{i=1}^{M}$, where $\boldsymbol{x}_{i,0}$ is the clean skeleton feature of video $i$, $\boldsymbol{v}_i$ is its aligned RGB feature, and $y_i\in\mathcal{Y}_s$ is the class label. For every candidate class $c\in\mathcal{Y}_s\cup\mathcal{Y}_u$, a frozen text encoder provides a local token matrix $\boldsymbol{T}_c$ and a global vector $\boldsymbol{g}_c$. At test time, the model assigns $(\boldsymbol{x}_{i,0},\boldsymbol{v}_i)$ to one class in $\mathcal{Y}_u$.

\paragraph{Core workflow.} Figure~\ref{fig:arch} illustrates the three modality paths and the shared scoring procedure used for training and inference. Following TDSM~\citep{do2025bridging}, we add noise only to $\boldsymbol{x}_{i,0}$ and do not sample a reverse diffusion chain for classification. The denoiser predicts the added noise from the noised skeleton under candidate-text and current-video RGB conditions, and the resulting prediction error serves as the candidate energy. Building on TDSM's candidate-wise diffusion scorer, TDSM-MM introduces RGB conditioning through a non-diffused RGB condition token that interacts jointly with the noised skeleton and each candidate text. During training, correct and wrong texts are evaluated with identical skeleton, RGB, timestep, and noise. During inference, the same quantities are fixed for each noise draw while unseen candidate text $c$ is varied, and the class with the lowest averaged energy is selected.

 \paragraph{Why joint conditioning instead of late fusion?} Independent late fusion assigns two scores to each candidate class $c$: a skeleton--text score $s_{\mathrm{skel}}(c)$ and an RGB--text score $s_{\mathrm{RGB}}(c)$, with lower values indicating better matches. For the true class $y_i$ and a competitor $c$, define the modality-specific margin as $\Delta_m(y_i,c)=s_m(c)-s_m(y_i)$ for $m\in\{\mathrm{skel},\mathrm{RGB}\}$. A fused score with skeleton weight $\alpha\in[0,1]$ ranks $y_i$ ahead of $c$ only if
\begin{equation}
    \alpha\Delta_{\mathrm{skel}}(y_i,c)
    +(1-\alpha)\Delta_{\mathrm{RGB}}(y_i,c)>0.
    \label{eq:margin}
\end{equation}
When the modalities disagree, the suitable range of $\alpha$ depends on the class pair and score scales, so a weight selected on seen classes may transfer unevenly to unseen pairs. TDSM-MM instead produces one energy conditioned on the skeleton, current RGB observation, and candidate text, introducing neither an independent RGB ranking nor an explicit post-hoc fusion weight.

\subsection{Multimodal Candidate-Conditioned Denoiser}

\paragraph{Frozen input representations.} A frozen Shift-GCN~\citep{cheng2020skeleton} maps each skeleton sequence to $\boldsymbol{x}_{i,0}\in\mathbb{R}^{d_s}$, where $d_s=256$. For candidate class $c$, we use the frozen OpenCLIP ViT-H text encoder employed by Stable Diffusion 2.1~\citep{cherti2023reproducible,rombach2022high} to encode two prompts separately: the action name and the fixed TDSM description. Both sequences are padded or truncated to $L=35$ positions. Concatenating the two token sequences along the feature dimension gives $\boldsymbol{T}_c\in\mathbb{R}^{L\times d_t}$, and concatenating their pooled representations gives $\boldsymbol{g}_c\in\mathbb{R}^{d_t}$, with $d_t=2048$. A frozen OpenCLIP ViT-H/14 visual encoder~\citep{cherti2023reproducible} processes eight uniformly sampled RGB frames. Each $1024$-dimensional frame feature is $\ell_2$-normalized; the features are then averaged and normalized again to obtain $\boldsymbol{v}_i\in\mathbb{R}^{d_v}$, where $d_v=1024$. All three encoders are frozen; the CrossDiT denoiser, including its input projections and learned positional embeddings, is optimized.

\paragraph{Skeleton corruption and token construction.} Following the standard DDPM forward process~\citep{ho2020denoising}, for timestep $t\in\{0,\ldots,T-1\}$ we corrupt only the clean skeleton feature:
\begin{equation}
  \boldsymbol{x}_{i,t}
  =\sqrt{\bar\alpha_t}\,\boldsymbol{x}_{i,0}
  +\sqrt{1-\bar\alpha_t}\,\boldsymbol{\epsilon}_i,
  \qquad
  \boldsymbol{\epsilon}_i\sim\mathcal{N}(\boldsymbol{0},\boldsymbol{I}),
  \label{eq:forward-diffusion}
\end{equation}
where $\bar\alpha_t$ controls the retained skeleton signal. The noised feature $\boldsymbol{x}_{i,t}\in\mathbb{R}^{d_s}$ and each row of $\boldsymbol{T}_c\in\mathbb{R}^{L\times d_t}$ are separately projected to the denoiser width $d=768$, producing one skeleton token and $L$ text tokens with learned positional embeddings. A learned affine mapping $E_v:\mathbb{R}^{d_v}\rightarrow\mathbb{R}^{d}$ and RGB slot embedding $\boldsymbol{p}_v$ produce one non-diffused RGB condition token:
\begin{equation}
  \boldsymbol{z}_i^{v}=E_v(\boldsymbol{v}_i)+\boldsymbol{p}_v\in\mathbb{R}^{d}.
  \label{eq:rgb-token}
\end{equation}
Because temporal order is removed by frame averaging, $\boldsymbol{v}_i$ mainly represents sample-level appearance and context, whereas $\boldsymbol{x}_{i,0}$ encodes temporal body dynamics.

\paragraph{Joint CrossDiT processing.} We append the RGB condition token $\boldsymbol{z}_i^{v}$ to the $L$ projected text tokens, forming the conditioning stream. Each CrossDiT block, inherited from TDSM and built on DiT~\citep{do2025bridging,peebles2023scalable}, uses separate query, key, value, and output projections for the skeleton and conditioning streams, followed by one joint multi-head attention operation. The outputs return to their respective streams and pass through separate feed-forward networks. Thus, every block processes $1+L+1=37$ tokens: one noised skeleton token, 35 text tokens, and one RGB condition token. The global text vector $\boldsymbol{g}_c$ is embedded and added to the timestep embedding to control AdaLN in both streams and in the final prediction layer. RGB does not generate AdaLN parameters; as part of the conditioning stream, however, its hidden state is normalized, text--time modulated, and updated by every block. The final layer maps the updated skeleton token to the predicted noise in $\mathbb{R}^{d_s}$.

\paragraph{Operational definition of the visual anchor.} We call $\boldsymbol{z}_i^{v}$ \emph{non-diffused} because it receives no forward-process corruption at the denoiser input. The same input RGB condition token is reused when candidate class $c$ changes, providing a sample-specific visual reference for every candidate evaluation. ``Stable'' refers only to this uncorrupted input: the token's hidden state still evolves inside CrossDiT and is modulated by candidate text and timestep. Consequently, RGB can modify the noise prediction differently under different class hypotheses without producing class logits or a separate RGB ranking.

\subsection{Paired Candidate Training}

\paragraph{Candidate denoising energy.} To keep notation compact, let $D_\theta(\boldsymbol{x}_{i,t},t;c,\boldsymbol{v}_i)$ denote the noise predicted by the denoiser when candidate $c$ supplies $(\boldsymbol{T}_c,\boldsymbol{g}_c)$. For each training sample, we draw $\boldsymbol{\epsilon}_i\sim\mathcal{N}(\boldsymbol{0},\boldsymbol{I})$ and $t_i$ uniformly from $\{0,\ldots,T-1\}$, then construct $\boldsymbol{x}_{i,t_i}$ with Eq.~\eqref{eq:forward-diffusion}. The candidate energy is the mean squared noise-prediction error over the $d_s$ skeleton-feature dimensions:
\begin{equation}
  \mathcal{E}_i(c)
  =\frac{1}{d_s}
  \left\|
  D_\theta(\boldsymbol{x}_{i,t_i},t_i;c,\boldsymbol{v}_i)
  -\boldsymbol{\epsilon}_i
  \right\|_2^2.
  \label{eq:sample-error}
\end{equation}

\paragraph{Correct- and wrong-text objective.} For each true label $y_i$, we sample $y_i^-\sim\operatorname{Uniform}(\mathcal{Y}_s)$. The two denoiser evaluations $\mathcal{E}_i(y_i)$ and $\mathcal{E}_i(y_i^-)$ share $\boldsymbol{x}_{i,0}$, $\boldsymbol{v}_i$, $t_i$, and $\boldsymbol{\epsilon}_i$; only the class text changes. To mask accidental matches, let $a_i=1$ if $y_i^-\neq y_i$ and $a_i=0$ otherwise. For a mini-batch of size $B$, the objective is
\begin{equation}
\begin{aligned}
  \mathcal{L}_{\mathrm{diff}}
  &=\frac{1}{B}\sum_{i=1}^{B}\mathcal{E}_i(y_i),\\
  \mathcal{L}_{\mathrm{TD}}
  &=\frac{1}{B}\sum_{i=1}^{B}a_i
    \big[\mathcal{E}_i(y_i)-\mathcal{E}_i(y_i^-)+\gamma\big]_+,\\
  \mathcal{L}
  &=\mathcal{L}_{\mathrm{diff}}+\mathcal{L}_{\mathrm{TD}}.
\end{aligned}
\label{eq:training-objective}
\end{equation}
where $[z]_+=\max(z,0)$. We use $\gamma=1$ and unit weights for both terms. The objective therefore trains one denoiser to assign lower energy to the correct class text than to a wrong class text under identical skeleton and visual evidence. It introduces no RGB classification loss, RGB-specific ranking loss, or comparison between different RGB samples.

\subsection{Candidate-Wise Zero-Shot Inference}

\paragraph{Scoring every unseen class.} We use one fixed timestep $t^\star$ and a fixed bank of $N$ independently sampled Gaussian noise vectors $\{\boldsymbol{\epsilon}^{(n)}\}_{n=1}^{N}$, shared across test samples. For a test sample $(\boldsymbol{x}_{i,0},\boldsymbol{v}_i)$ and each noise draw, we construct $\boldsymbol{x}_{i,t^\star}^{(n)}$ once using Eq.~\eqref{eq:forward-diffusion}, then reuse the same noised skeleton, RGB feature, timestep, and noise while evaluating every candidate $c\in\mathcal{Y}_u$. The candidate energy, averaged score, and prediction are
\begin{equation}
\begin{aligned}
  \mathcal{E}_{i,n}(c)
  &=\frac{1}{d_s}
    \left\|
    D_\theta(\boldsymbol{x}_{i,t^\star}^{(n)},t^\star;c,\boldsymbol{v}_i)
    -\boldsymbol{\epsilon}^{(n)}
    \right\|_2^2,\\
  S_i(c)
  &=\frac{1}{N}\sum_{n=1}^{N}\mathcal{E}_{i,n}(c),\\
  \widehat{y}_i
  &=\arg\min_{c\in\mathcal{Y}_u}S_i(c).
\end{aligned}
\label{eq:mm-inference}
\end{equation}
Only candidate text changes inside the class loop; the current sample's RGB feature never supplies an independent ranking. After checkpoint selection, inference updates neither model parameters nor prompts, prototypes, or caches from the incoming unseen test stream. The experimental setup specifies $t^\star$ and $N$.

\section{Experiments}
\label{sec:experiments}
\subsection{Experimental Setup}
\paragraph{Datasets and splits.} We evaluate on NTU RGB+D 60~\citep{shahroudy2016ntu} and NTU RGB+D 120~\citep{liu2019ntu} under the standard SynSE~\citep{gupta2021syntactically} seen/unseen splits (48/12 and 55/5 for NTU-60; 96/24 and 110/10 for NTU-120). Following the primary TDSM protocol, models are trained using only seen-class samples, and we report top-1 zero-shot accuracy on unseen classes. After the inherited checkpoint-selection procedure, inference uses a fixed model without updating parameters, prompts, prototypes, or caches from the incoming target stream.
  
\paragraph{Features and backbones.} Following SynSE, skeleton features are extracted using a frozen Shift-GCN~\citep{cheng2020skeleton} (256-d). RGB features are extracted using a frozen OpenCLIP ViT-H/14 visual encoder~\citep{cherti2023reproducible} (1024-d). Class semantic embeddings are extracted using the frozen OpenCLIP ViT-H text encoder employed by Stable Diffusion 2.1~\citep{cherti2023reproducible,rombach2022high}, with the original TDSM~\citep{do2025bridging}  class-name and LLM-description prompts.

\paragraph{Comparison protocol.} We use the published skeleton-only TDSM~\citep{do2025bridging} as the closest paradigm reference, sharing the frozen skeleton/text representations and CrossDiT design. Our content-free constant-slot ablation provides the same-architecture control for sample-specific RGB content. Flora is the strongest external inductive baseline, while DynaPURLS is a representative transductive baseline.

\paragraph{Implementation.} Our denoiser is a 12-block transformer~\citep{peebles2023scalable}
with hidden dimension 768, trained from scratch. We optimize it for 50k iterations with AdamW, a learning rate of $10^{-4}$, weight decay of $0.01$, a 100-step warmup, cosine scheduling, and batch size 256. Training uses a 50-step DDPM scheduler with $\epsilon$-prediction. At inference, we follow TDSM and use timestep $t=25$ with $N=10$ noise samples for denoising-error estimation. Unless noted, reported accuracies are averaged over five evaluation runs with independent noise samples.

\subsection{Comparison with State-of-the-Art Methods.}
Table~\ref{tab:main} compares TDSM-MM with representative zero-shot action
recognition methods, including skeleton-only approaches, the diffusion baseline
TDSM~\citep{do2025bridging}, the RGB-enhanced BSZSL~\citep{liu2025beyond},
and recent inductive and transductive methods.

Compared with skeleton-only TDSM, TDSM-MM improves accuracy on all four SynSE
splits, with the largest gains on NTU-60 48/12, NTU-120 96/24, and NTU-120
110/10 ($+10.2$, $+6.2$, and $+12.4$ percentage points, respectively). This
demonstrates that RGB provides complementary information beyond skeleton
dynamics, while the limited gain on NTU-60 55/5 is consistent with its near
saturation. The improvement over BSZSL~\citep{liu2025beyond} and the weak
performance of RGB-only OpenCLIP matching ($19.3\%$--$60.3\%$) further indicate
that the gain comes from effective skeleton--RGB interaction rather than RGB
features alone.

Among inductive methods, TDSM-MM achieves the best performance on three of four
splits, outperforming Flora~\citep{chen2026learning} by $0.9$, $4.9$, and
$7.0$ percentage points on NTU-60 48/12, NTU-120 96/24, and NTU-120 110/10,
respectively. Notably, on NTU-120 96/24, TDSM-MM reaches $71.3\%$ and exceeds
the transductive DynaPURLS~\citep{zhu2026dynapurls} ($69.1\%$) without using
target-stream statistics or test-time adaptation.

\begin{table*}[t]
\centering
\small
\begin{tabular}{llcccccc}
\toprule
\multirow{2}{*}{\textbf{Method}} &
\multirow{2}{*}{\textbf{Paradigm}} &
\multirow{2}{*}{\textbf{Modality}} &
\multirow{2}{*}{\textbf{Setting}} &
\multicolumn{2}{c}{\textbf{NTU-60}} &
\multicolumn{2}{c}{\textbf{NTU-120}} \\
\cmidrule(lr){5-6} \cmidrule(lr){7-8}
  & & & & \textbf{48/12} & \textbf{55/5} & \textbf{96/24} & \textbf{110/10} \\
\midrule
\multicolumn{8}{l}{\emph{Skeleton-only, inductive}}\\
SynSE~\citep{gupta2021syntactically} & alignment & S & inductive & 33.3 & 75.8 & 38.7 & 62.7 \\
SMIE~\citep{zhou2023zero} & alignment & S & inductive & 40.2 & 78.0 & 45.3 & 65.7 \\
SA-DVAE~\citep{li2024sa} & VAE & S & inductive & 41.4 & 82.4 & 46.1 & 68.8 \\
PURLS~\citep{zhu2024part} & alignment & S & inductive & 41.0 & 79.2 & 52.0 & 72.0 \\
FS-VAE~\citep{wu2025frequency} & VAE & S & inductive & 57.2 & 86.9 & 62.5 & 74.4 \\
Neuron~\citep{chen2025neuron} & prototype & S & inductive & 62.7 & 86.9 & 57.1 & 71.5 \\
SkeletonContext~\citep{wang2026skeletoncontext} & alignment+ctx & S & inductive & 64.4 & \textbf{89.6} & 60.1 & 74.2 \\
TDSM~\citep{do2025bridging} & diffusion & S & inductive & 56.0 & 86.5 & 65.1 & 74.2 \\
Flora$^\dagger$~\citep{chen2026learning} & flow & S & inductive & \textbf{65.3} & 86.3 & \textbf{66.4} & \textbf{79.6} \\
\midrule
\multicolumn{8}{l}{\emph{Transductive (uses test-time statistics)}}\\
DynaPURLS~\citep{zhu2026dynapurls} & TTA & S & \emph{transductive} & 71.8 & 88.5 & 69.1 & 89.1 \\
\midrule
\multicolumn{8}{l}{\emph{RGB-only (appearance matching), training-free}}\\
OpenCLIP ViT-H/14~\citep{cherti2023reproducible} & matching & R & inductive & 19.3 & 60.3 & 18.5 & 41.3 \\
\midrule
\multicolumn{8}{l}{\emph{multimodal, inductive}}\\
BSZSL~\citep{liu2025beyond} & alignment & S+R & inductive & 53.0 & 83.0 & 56.1 & 77.7 \\
\textbf{TDSM-MM (ours)} & diffusion & S+R & inductive & \textbf{66.2} & \textbf{86.8} & \textbf{71.3} & \textbf{86.6} \\
\bottomrule
\end{tabular}
\caption{Top-1 ZSL accuracy (\%) on the NTU-60/120 SynSE splits (S=skeleton, R=RGB). Boldface entries mark the best inductive result within each input-modality group. DynaPURLS uses target-stream statistics and is shown for reference; $^\dagger$ denotes Flora's reported 4s-Shift-GCN result.}
\label{tab:main}
\end{table*}

\begin{table}[t]
  \centering
  \small
  \setlength{\tabcolsep}{1mm}
  \begin{tabular}{@{}llc@{}}
  \toprule
  \textbf{Variant} & \textbf{RGB interface} & \textbf{Acc. (\%)} \\
  \midrule
  \multicolumn{3}{l}{\emph{Diffusion routing/training controls}}\\
  Constant slot & content-free condition token & 56.5 \\
  Skeleton-side RGB & skeleton-side early fusion & 59.4 \\
  RGB-corrupted training & noised RGB condition token & 56.3 \\
  TDSM-MM & non-diffused condition token & \textbf{66.2} \\
  \midrule
  \multicolumn{3}{l}{\emph{Matched discriminative CrossDiT controls}}\\
  D0: constant slot & content-free condition token & 49.6 \\
  D1: skeleton-side RGB & skeleton-side early fusion & 48.2 \\
  D2: RGB conditioning & non-diffused condition token & \textbf{50.9} \\
  \bottomrule
  \end{tabular}
  \caption{Ablations on NTU-60 48/12. The noised-token variant corrupts RGB only during training and uses non-diffused RGB at inference. In D1, a content-free condition token slot is retained for capacity matching, while sample-specific RGB enters through skeleton-side early fusion.}
  \label{tab:rgb_ablation}
  \end{table}

\subsection{Ablations}

\paragraph{Effect of RGB conditioning design.}
Table~\ref{tab:rgb_ablation} evaluates different variants of the RGB conditioning mechanism, including content removal, alternative fusion, and condition corruption. Removing sample-specific RGB information reduces accuracy from $66.2\%$ to $56.5\%$, showing that the improvement comes from informative visual evidence rather than merely introducing an additional condition token slot. Injecting RGB features into the skeleton-side token achieves only $59.4\%$, while corrupting the RGB condition token reduces performance to $56.3\%$. These comparisons show that, among the tested designs, a separate and non-diffused RGB condition token provides the most effective guidance for diffusion-based recognition.

\paragraph{Interaction with the scoring formulation.}
To test whether RGB conditioning benefits specifically from diffusion-based scoring, we compare matched discriminative and diffusion CrossDiT variants. Adding the RGB condition token improves the discriminative model from $49.60\%$ to $50.86\%$, but improves the diffusion model from $56.94\%$ to $66.67\%$. The gain difference is $8.47$ percentage points, with a 95\% class-stratified bootstrap confidence interval of $[6.39,10.55]$, indicating that denoising-based scoring substantially amplifies RGB conditioning.

\subsection{Mechanism Analysis}

\paragraph{RGB conditioning versus score fusion.}
We freeze a constant-slot skeleton-diffusion scorer and a seen-trained RGB--text scorer, then fit temperature-calibrated global fusion and select a two-layer candidate-wise gate over score, confidence, and rank features on four seen-class folds. On the same ten-draw bank, the skeleton scorer, calibrated fusion, learned gate, and joint conditioning reach $56.94\%$, $60.52\%$, $57.03\%$, and $66.67\%$, respectively. The strongest tested independent fusion remains $6.15$ percentage points below joint conditioning, with a paired class-stratified 95\% bootstrap confidence interval of $[4.68,7.58]$.

\paragraph{Candidate-dependent RGB influence.} We first perform a paired counterfactual intervention at the default timestep $t=25$: the checkpoint, skeleton, candidate texts, timestep, and ten-noise bank are fixed, while only the RGB feature is replaced. Matched, same-action, and controlled different-action RGB yield $66.39\%$, $66.06\%$, and $53.12\%$ accuracy, respectively. Under different-action replacement, the true-versus-donor margin shifts toward the donor class for $78.0\%$ of samples, the donor improves by an average of $1.22$ ranking positions, and $7.8\%$ of predictions turn to the donor. Thus, RGB changes relative candidate energies rather than contributing only a candidate-independent offset.

  \begin{figure}[t]
  \centering
  \includegraphics[width=\columnwidth]{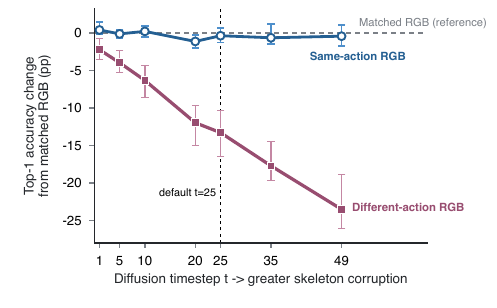}
  \caption{Effect of RGB substitution across diffusion timesteps on NTU-60 48/12. Action-incompatible RGB becomes increasingly harmful as skeleton corruption grows, whereas same-action RGB remains close
  to matched RGB.}
  \label{fig:anchor_timestep}
  \end{figure}

\paragraph{Corruption-dependent visual anchoring.}
Replacing matched RGB with different-action RGB reduces accuracy by $2.20$ and $23.52$ percentage points at $t=1$ and $t=49$, respectively. The paired endpoint change is $-21.31$ percentage points, with a subject-cluster 95\% bootstrap confidence interval of $[-23.52,-17.32]$, while same-action replacement remains stable. This timestep-dependent effect is consistent with RGB serving as a visual anchor when skeleton evidence becomes increasingly corrupted.

\paragraph{Reliability of RGB conditioning.}
A $2{\times}2$ clean/noisy RGB factorial experiment further separates training-time condition reliability from test-time sensitivity. Test-time RGB corruption reduces the clean-trained model from $66.39\%$ to $64.92\%$, but has little effect on the noisy-trained model ($55.32\%\rightarrow55.23\%$). The top-1 interaction is $1.38$ percentage points (95\% bootstrap confidence interval $[-0.06,2.88]$), while the candidate-margin interaction is $0.0343$ (95\% bootstrap confidence interval $[0.0238,0.0446]$), suggesting that clean-condition training increases sensitivity to RGB reliability.

\begin{table}[t]
  \centering
  \small
  
  \begin{tabular}{@{}lc@{}}
  \toprule
  \textbf{RGB input} & \textbf{Top-1 (\%)} \\
  \midrule
  Matched current-video RGB & \textbf{86.6} \\
  Zero vector & 52.6 \\
  Seen-training mean & 40.5 \\
  Label-mismatched real RGB & 35.5 \\
  \bottomrule
  \end{tabular}
  \caption{Paired RGB-content interventions on NTU-120 110/10 using a fixed checkpoint and ten shared noise draws. Only the RGB input changes.}
  \label{tab:rgb_content_intervention}
\end{table}

\paragraph{Sample-specific visual evidence.}
To verify that RGB conditioning exploits instance-level visual cues rather than only an RGB pathway, we replace the original RGB feature with zero, mean, or label-mismatched RGB while keeping the skeleton, text, and checkpoint fixed. Table~\ref{tab:rgb_content_intervention} shows that matched current-video RGB substantially outperforms all alternatives, indicating that the gain comes from sample-specific visual evidence rather than a generic visual prior. The effect is class-dependent: visually ambiguous classes retain larger gains, whereas some motion-dominant classes show limited or negative changes, suggesting that RGB conditioning mainly reshapes competition among visually confusable actions.

\section{Discussion and Limitations}
\paragraph{When RGB conditioning is effective.}
RGB conditioning is most beneficial when visual information provides complementary evidence beyond skeleton motion. The consistent improvements on NTU-60 48/12 and NTU-120 96/24 and 110/10 ($+10.2$, $+6.2$, and $+12.4$ percentage points over skeleton-only TDSM, respectively) suggest that conditioning is particularly useful when the skeleton--text correspondence is insufficient to resolve unseen classes. On NTU-120 96/24, TDSM-MM further surpasses the transductive DynaPURLS ($71.3\%$ vs.\ $69.1\%$) without using any test-time statistics, trading additional
per-sample RGB observations for a fully inductive inference procedure.

\paragraph{Modality fairness.}
Our matched reference is the same-backbone, same-paradigm skeleton-only TDSM
($\pm$RGB); under this controlled comparison, adding RGB consistently improves
aggregate accuracy on three of four splits and is near-neutral on the fourth,
reversing the prior negative evidence (BSZSL).

\paragraph{Limitations and future directions.} RGB conditioning depends on informative and correctly aligned visual observations: label-mismatched real RGB reduces accuracy from $86.6\%$ to $35.5\%$ on the fixed NTU-120 110/10 checkpoint, while \emph{salute} regresses relative to all three reference conditions. The corruption-dependent and per-class analyses characterize fixed checkpoints and therefore quantify test-population rather than training-seed uncertainty. Inference requires $O(K{\cdot}N)$ denoiser evaluations per sample; with cached RGB features, the additional RGB condition token increases denoiser parameters by $0.30\%$ and measured scorer latency by $3.08$--$4.37\%$, while online deployment additionally incurs the frozen ViT-H/14 encoding cost. Handling missing, unreliable, or misleading RGB remains an important direction.

\section{Conclusion}
In this paper, we showed that in inductive ZSAR, supplying RGB information as conditioning of a single denoising objective is more effective than combining independent modality scores. Score combination is limited by the absence of unseen-class supervision for weight calibration; conditioning avoids the weighting problem by absorbing fusion into learned attention.
TDSM-MM instantiates this principle with a single non-diffused RGB condition token, which serves as a stable visual anchor during diffusion denoising. It achieves the best inductive accuracy on three of four NTU-60/120 splits, and on 96/24 surpasses the transductive state of the art without using any test-time statistics. However, per-class competition remains: conditioning mitigates but does not eliminate regressions on visually confusable classes, and the gain depends on RGB carrying class-relevant cues that skeletons lack. Addressing these residual regressions is a direction for future work.

\FloatBarrier
{\small
\bibliographystyle{ieeenat_fullname}
\bibliography{references}
}

\clearpage
\input{appendix}

\end{document}

%% file: appendix.tex
\appendix
\setcounter{table}{0}
\setcounter{figure}{0}
\setcounter{equation}{0}
\setcounter{algorithm}{0}
\renewcommand{\thetable}{S\arabic{table}}
\renewcommand{\thefigure}{S\arabic{figure}}
\renewcommand{\theequation}{S\arabic{equation}}
\renewcommand{\thealgorithm}{S\arabic{algorithm}}
\renewcommand{\theHtable}{S\arabic{table}}
\renewcommand{\theHfigure}{S\arabic{figure}}
\renewcommand{\theHequation}{S\arabic{equation}}
\renewcommand{\theHalgorithm}{S\arabic{algorithm}}

\fullwidthtitle{Technical Supplement}

This technical supplement provides implementation and evaluation details, expanded controls for RGB routing, the scoring objective, and independent-score fusion, additional interventions characterizing non-diffused RGB conditioning, and sensitivity and efficiency analyses. The main paper is self-contained; the material below provides extended results and protocol-level auditability.

\section{Implementation and Evaluation Protocol}

\subsection{Datasets and Zero-Shot Protocols}

We evaluate on the NTU RGB+D 60 and NTU RGB+D 120 datasets~\citep{shahroudy2016ntu,liu2019ntu} using the four class partitions adopted by TDSM~\citep{do2025bridging}. Table~\ref{tab:dataset_protocols} lists the number of seen-class training samples and unseen-class test samples after applying the corresponding cross-subject and class filters. Training uses paired skeleton and RGB observations from seen classes, whereas candidate-wise evaluation is restricted to the designated unseen classes.

\noindent\begin{minipage}{\columnwidth}
\centering
\small
\begin{tabular}{lrrr}
\toprule
Protocol & Seen/unseen & Train & Test \\
\midrule
NTU-60 48/12 & 48/12 & 31,882 & 3,270 \\
NTU-60 55/5 & 55/5 & 36,548 & 1,356 \\
NTU-120 96/24 & 96/24 & 50,116 & 9,883 \\
NTU-120 110/10 & 110/10 & 57,469 & 3,948 \\
\bottomrule
\end{tabular}
\normalsize
\captionsetup{hypcap=false}
\captionof{table}{Dataset protocols used in the main experiments. ``Train'' contains labeled seen-class samples; ``test'' contains samples from the designated unseen classes.}
\label{tab:dataset_protocols}
\end{minipage}

We follow the checkpoint-selection and reporting convention of the original TDSM benchmark and, for reproducibility, use a fixed training seed of 2027. The inherited checkpoint-evaluation interface uses three diffusion-noise draws during training; after selection, the checkpoint is kept fixed, and the reported main results are re-evaluated with ten draws. Candidate scoring does not update model parameters, prompts, prototypes, or caches from incoming target samples, and all counterfactual analyses reuse the same fixed checkpoint within each comparison.

\subsection{Frozen Input Representations}

\paragraph{Skeleton.} A frozen Shift-GCN~\citep{cheng2020skeleton} maps each skeleton sequence to $\boldsymbol{x}_{i,0}\in\mathbb{R}^{d_s}$, where $d_s=256$. This is the only modality subjected to the DDPM forward process~\citep{ho2020denoising}.

\paragraph{Text.} We use the frozen OpenCLIP-compatible ViT-H text encoder employed by Stable Diffusion 2.1~\citep{cherti2023reproducible,rombach2022high}. The action name and the fixed TDSM description are encoded separately, each padded or truncated to 35 positions. Concatenating the two last-hidden-state sequences along the channel dimension gives $\boldsymbol{T}_c\in\mathbb{R}^{35\times2048}$ for candidate $c$; concatenating their pooled outputs gives $\boldsymbol{g}_c\in\mathbb{R}^{2048}$.

\paragraph{RGB.} A frozen OpenCLIP ViT-H/14 visual encoder pretrained on LAION-2B~\citep{cherti2023reproducible} processes eight frames sampled uniformly over video $i$. Each 1024-dimensional frame embedding is $\ell_2$-normalized, the embeddings are averaged over time, and the average is normalized again. A learned linear layer maps the resulting $\boldsymbol{v}_i\in\mathbb{R}^{1024}$ to one 768-dimensional RGB condition token.

\begin{table*}[t]
\centering
\small
\begin{tabular}{llll}
\toprule
Quantity & Native size & CrossDiT size & Role \\
\midrule
Skeleton feature $\boldsymbol{x}_{i,0}$ & $256$ & $1\times768$ & Diffused target representation \\
Local text $\boldsymbol{T}_c$ & $35\times2048$ & $35\times768$ & Candidate-specific condition tokens \\
Global text $\boldsymbol{g}_c$ & $2048$ & $768$ & AdaLN modulation with timestep \\
Framewise RGB & $8\times1024$ & --- & Uniformly sampled visual observations \\
Pooled RGB $\boldsymbol{v}_i$ & $1024$ & $1\times768$ & Non-diffused RGB condition token \\
Joint attention sequence & --- & $37\times768$ & 1 skeleton, 35 text, and 1 RGB condition token \\
\bottomrule
\end{tabular}
\normalsize
\caption{Representations and tensor sizes. All three feature encoders are frozen; the projections, positional embeddings, CrossDiT blocks, and output layer are trainable.}
\label{tab:tensor_sizes}
\end{table*}

\subsection{Denoiser Architecture}

We retain the main-paper notation $D_\theta$ for the denoiser, which inherits the 12-block CrossDiT design of TDSM, with width 768, 12 attention heads, and an MLP expansion ratio of 4. Each block first forms separate query, key, and value projections for the skeleton stream and the conditioning stream, concatenates the projected tokens for joint attention, and then splits the two streams before separate feed-forward networks. The sum of the timestep embedding and projected global candidate-text embedding controls AdaLN modulation in both streams. RGB does not independently control AdaLN and does not produce class logits; it affects the predicted skeleton noise through joint attention with the candidate text and noised skeleton.

The term \emph{non-diffused} refers specifically to the input route: the RGB feature is not corrupted by the forward diffusion process. Its hidden representation still changes through AdaLN, attention, and feed-forward updates inside CrossDiT. The reported model uses one RGB condition token, RGB condition dropout probability zero, and guidance scale one.

\subsection{Training and Candidate-Wise Inference}

For a seen-class sample $(\boldsymbol{x}_{i,0},\boldsymbol{v}_i,y_i)$, training samples one timestep $t_i$ from the $T=50$ step schedule, one Gaussian noise vector $\boldsymbol{\epsilon}_i$, and one wrong seen-class label $y_i^-$. The correct and wrong passes share $\boldsymbol{x}_{i,t_i}$, $\boldsymbol{v}_i$, $t_i$, and $\boldsymbol{\epsilon}_i$; only the candidate text changes. The implementation samples $y_i^-$ uniformly from the complete seen label set and masks the triplet term when $y_i^-=y_i$. Algorithm~\ref{alg:training} summarizes one training update.

\begin{algorithm}[!ht]
\caption{TDSM-MM training update}
\label{alg:training}
\raggedright
\textbf{Input}: Seen sample $(\boldsymbol{x}_{i,0},\boldsymbol{v}_i,y_i)$, seen label set $\mathcal{Y}_s$, and frozen class-text embeddings\\
\textbf{Output}: Updated denoiser parameters $\theta$
\begin{algorithmic}[1]
\STATE Sample $t_i\sim\mathrm{Uniform}\{0,\ldots,T-1\}$
\STATE Sample $\boldsymbol{\epsilon}_i\sim\mathcal{N}(\boldsymbol{0},\boldsymbol{I})$
\STATE $\boldsymbol{x}_{i,t_i}\gets\sqrt{\bar{\alpha}_{t_i}}\boldsymbol{x}_{i,0}+\sqrt{1-\bar{\alpha}_{t_i}}\boldsymbol{\epsilon}_i$
\STATE Set $y_i^+\gets y_i$ and sample $y_i^-\sim\mathrm{Uniform}(\mathcal{Y}_s)$
\FOR{$q\in\{+,-\}$}
\STATE $\hat{\boldsymbol{\epsilon}}_i^q\gets D_\theta(\boldsymbol{x}_{i,t_i},t_i;y_i^q,\boldsymbol{v}_i)$
\STATE $\mathcal{E}_i(y_i^q)\gets d_s^{-1}\lVert\hat{\boldsymbol{\epsilon}}_i^q-\boldsymbol{\epsilon}_i\rVert_2^2$
\ENDFOR
\STATE $\ell_i\gets\max\!\left(0,\mathcal{E}_i(y_i)-\mathcal{E}_i(y_i^-)+\gamma\right)$
\STATE $\mathcal{L}_i\gets\mathcal{E}_i(y_i)+\mathbf{1}[y_i^-\ne y_i]\ell_i$
\STATE Update $\theta$ using the batch mean of $\mathcal{L}_i$
\end{algorithmic}
\end{algorithm}

At inference, the model uses a fixed timestep $t^\star=25$ and a bank of $N=10$ Gaussian noise draws. For each draw, the noised skeleton and RGB are held fixed while candidate text is substituted. The candidate score is the mean squared error (MSE) of the noise prediction averaged over the shared bank, and the minimum-score candidate is selected. Algorithm~\ref{alg:inference} summarizes this procedure, which requires $K N$ denoiser evaluations for $K$ unseen candidates and does not compute an independent RGB ranking.

\begin{algorithm}[!ht]
\caption{Candidate-wise zero-shot inference}
\label{alg:inference}
\raggedright
\textbf{Input}: Sample $(\boldsymbol{x}_{i,0},\boldsymbol{v}_i)$, unseen candidates $\mathcal{Y}_u$, timestep $t^\star$, and shared noise bank $\{\boldsymbol{\epsilon}^{(n)}\}_{n=1}^{N}$\\
\textbf{Output}: Predicted unseen class $\widehat{y}_i$
\begin{algorithmic}[1]
\FOR{$n=1,\ldots,N$}
\STATE $\boldsymbol{x}_{i,t^\star}^{(n)}\gets\sqrt{\bar{\alpha}_{t^\star}}\boldsymbol{x}_{i,0}+\sqrt{1-\bar{\alpha}_{t^\star}}\boldsymbol{\epsilon}^{(n)}$
\FOR{each $c\in\mathcal{Y}_u$}
\STATE $\widehat{\boldsymbol{\epsilon}}_{i,n}(c)\gets D_\theta(\boldsymbol{x}_{i,t^\star}^{(n)},t^\star;c,\boldsymbol{v}_i)$
\STATE $\mathcal{E}_{i,n}(c)\gets d_s^{-1}\lVert\widehat{\boldsymbol{\epsilon}}_{i,n}(c)-\boldsymbol{\epsilon}^{(n)}\rVert_2^2$
\ENDFOR
\ENDFOR
\FOR{each $c\in\mathcal{Y}_u$}
\STATE $S_i(c)\gets N^{-1}\sum_{n=1}^{N}\mathcal{E}_{i,n}(c)$
\ENDFOR
\STATE \textbf{return} $\widehat{y}_i\gets\arg\min_{c\in\mathcal{Y}_u}S_i(c)$
\end{algorithmic}
\end{algorithm}

\begin{table*}[t]
\centering
\small
\begin{tabular}{llll}
\toprule
Item & Value & Item & Value \\
\midrule
Optimizer & AdamW & Learning rate & $10^{-4}$ \\
Weight decay & 0.01 & LR schedule & 100-step warmup, cosine \\
Updates & 50,000 & Batch size & 256 \\
DDPM steps $T$ & 50 & Prediction target & $\boldsymbol{\epsilon}$ \\
Diffusion-loss weight & 1 & Triplet-loss weight & 1 \\
Triplet margin $\gamma$ & 1 & Wrong texts/update & 1 per sample \\
Inference timestep & 25 & Reported noise draws & 10 \\
Test batch size & 512 & Feature workers & 4 \\
Python / PyTorch & 3.9.19 / 2.4.0 & CUDA / cuDNN & 11.8 / 9.1 \\
Diffusers / Transformers & 0.32.2 / 4.44.2 & Accelerate / OpenCLIP & 0.33.0 / 3.3.0 \\
Hardware & RTX 4090, 24 GB & Frozen RGB frames & 8 per video \\
\bottomrule
\end{tabular}
\normalsize
\caption{Optimization, inference, and execution settings used by TDSM-MM.}
\label{tab:implementation}
\end{table*}

\subsection{Statistical Reporting}

Unless stated otherwise, each paired 95\% confidence interval (CI) is estimated using 10,000 class-stratified bootstrap replicates, resampling test examples within each true class while preserving the observed class counts. Paired accuracy comparisons additionally use exact two-sided McNemar tests. The NTU-120 110/10 analysis reports a subject-cluster bootstrap as a sensitivity check. Intervals from these procedures quantify paired test-population effects, while variation across diffusion-noise banks quantifies Monte Carlo evaluation variation.

\section{Expanded Controlled Comparisons}

\subsection{RGB Routing Controls}

Table~\ref{tab:routing_controls} expands the routing ablation from the main paper. The constant-slot model retains a condition-token position but removes sample-specific RGB content. The skeleton-side model projects RGB and adds it to the skeleton token before CrossDiT while keeping a content-free condition slot for capacity matching. The RGB-corrupted model applies the DDPM schedule to the RGB condition during training but evaluates with non-diffused RGB. All rows preserve the same candidate-wise diffusion-scoring interface.

\begin{table}[t]
\centering
\small
\begin{tabular}{lc}
\toprule
Diffusion route & Top-1 (\%) \\
\midrule
Content-free constant slot & $56.50\pm0.75$ \\
Skeleton-side RGB & $59.43\pm0.64$ \\
RGB-corrupted training & $56.28\pm0.39$ \\
\textbf{Non-diffused RGB condition token} & $\mathbf{66.18\pm0.68}$ \\
\bottomrule
\end{tabular}
\normalsize
\caption{Controlled RGB-routing variants on NTU-60 48/12. Values are means and standard deviations (SDs) over independently sampled evaluation-noise banks for each evaluated variant.}
\label{tab:routing_controls}
\end{table}

Replacing the constant slot with informative, non-diffused RGB improves accuracy by 9.68 percentage points. Skeleton-side routing retains part of the benefit but remains 6.75 points below the separate RGB condition, while corrupting the RGB condition during training reduces it to the constant-slot level. Within these controlled routes, both visual content and its non-diffused conditioning path are important.

\subsection{Matched Discriminative CrossDiT}

To separate the denoising objective from generic multimodal attention, we construct a discriminative CrossDiT with the same 12-block, 768-wide, 12-head architecture, identical frozen inputs, 37-token interface, optimizer, 50,000-update budget, batch order, wrong-label stream, and shared initialization across its three routing arms. Instead of predicting diffusion noise, it produces candidate scores and minimizes the pairwise ranking objective $\mathrm{softplus}(q_i^- - q_i^+)$ for one positive and one wrong candidate per sample. D0 uses a content-free token, D1 adds RGB to the skeleton-side token, and D2 uses an RGB condition token.

\begin{table}[t]
\centering
\small
\begin{tabular}{lrrr}
\toprule
Objective & Content-free & RGB condition & RGB gain \\
\midrule
Diffusion & 56.94 & 66.67 & $+9.72$ \\
Discriminative & 49.60 & 50.86 & $+1.25$ \\
\bottomrule
\end{tabular}
\normalsize
\caption{Objective-matched RGB control on NTU-60 48/12. The RGB gain compares the condition-token model with the corresponding content-free model.}
\label{tab:objective_control}
\end{table}

The diffusion RGB gain is $+9.72$ points (95\% CI $[8.26,11.22]$), whereas the discriminative gain is $+1.25$ points (95\% CI $[-0.24,2.78]$). Their interaction is $+8.47$ points with a class-stratified 95\% CI of $[6.39,10.55]$. Within the discriminative objective, D2 also exceeds skeleton-side D1 ($48.17\%$) by $2.69$ points (95\% CI $[1.16,4.25]$). The token route can therefore help a discriminative model, but the substantially larger gain emerges when RGB conditioning participates in candidate-wise denoising.

\subsection{Independent-Score Fusion and Learned Gating}

We next test whether a stronger independent-score rule can recover the joint scorer's gain. The skeleton branch is the constant-slot diffusion scorer, and the RGB branch is a two-layer projection head trained on the 48 seen classes against OpenCLIP text prototypes. We form 12-way seen-class episodes and use four gate-only class-held-out folds: both frozen base branches have seen all 48 training classes, while temperatures, fusion weights, gate features, and gate architecture are selected without unseen examples or scores. Each learned gate is a two-layer MLP. The sample-wise gate emits one skeleton weight per sample; the candidate-wise gate emits one weight per sample--candidate pair from calibrated scores, confidence summaries, and normalized ranks.

\begin{table}[t]
\centering
\small
\begin{tabular}{lc}
\toprule
Scoring route & Top-1 (\%) \\
\midrule
Skeleton diffusion score & 56.94 \\
Independent RGB score & 38.50 \\
Normalized fixed fusion & 56.94 \\
Temperature-calibrated fixed fusion & \textbf{60.52} \\
Entropy-conditioned rule & 60.12 \\
Margin-conditioned rule & 56.12 \\
Sample-wise learned gate & 54.62 \\
Candidate-wise learned gate & 57.03 \\
\midrule
\textbf{Joint diffusion scorer} & $\mathbf{66.67}$ \\
\bottomrule
\end{tabular}
\normalsize
\caption{Independent-score routes on NTU-60 48/12.}
\label{tab:learned_gating}
\end{table}

The best tested independent route is temperature-calibrated fixed fusion at $60.52\%$, which remains 6.15 points below joint conditioning (95\% CI $[4.68,7.58]$). The preselected candidate-wise learned gate reaches $57.03\%$; the joint scorer exceeds it by 9.63 points (95\% CI $[8.10,11.16]$, exact McNemar $p=1.41\times10^{-26}$). Thus, within the tested independent-score family, seen-trained adaptive fusion does not close the gap to a scorer in which RGB, skeleton, and candidate text interact before a single denoising energy is produced.

\paragraph{Correctness overlap.} We additionally compare the per-sample correctness sets of the constant-slot skeleton scorer (F0), independent RGB scorer, and joint diffusion scorer (F2) on the same 3,270 examples. F2 rescues 569 F0 errors, yielding a net gain of 318 correct predictions. More importantly, F0 and RGB are both wrong on 961 examples, yet F2 correctly classifies 386 of them (40.17\%; class-stratified 95\% CI $[37.26,43.00]$). These 386 predictions lie outside the union of the two independent branches' correct sets, showing that joint conditioning produces correct decisions unavailable from either tested independent top-1 output rather than merely selecting between them.

\section{Additional Analysis of Visual Anchoring}

\subsection{Counterfactual RGB Interventions}

For each test sample, we construct three RGB conditions while keeping the checkpoint, skeleton, candidate texts, timestep, and complete diffusion-noise bank fixed. \emph{Matched} uses the sample's own RGB. \emph{Same-action} matches setup, camera, performer, and action but uses the other repetition. \emph{Different-action} matches setup, camera, performer, and repetition but uses another action from the unseen candidate set; its donor label and RGB feature are denoted by $\widetilde{y}_i$ and $\boldsymbol{v}_i^{\mathrm{donor}}$, respectively. The donor-label mapping is deterministic, has no fixed points, and is constructed before evaluation; missing alternate-repetition features are extracted using the same frozen eight-frame RGB encoder.

Let $S_i(c;\boldsymbol{v})$ denote the candidate score averaged over the shared noise bank when class $c$ is evaluated under RGB feature $\boldsymbol{v}$, extending the score $S_i(c)$ defined in Algorithm~\ref{alg:inference}. We report the true-versus-best-wrong margin
\begin{equation}
m_i(\boldsymbol{v})=\min_{c\ne y_i}S_i(c;\boldsymbol{v})-S_i(y_i;\boldsymbol{v}),
\label{eq:classification_margin}
\end{equation}
which is positive when the true class outranks every alternative. For a different-action donor, we define its relative score gap $r_i(\boldsymbol{v})$ and the induced shift as
\begin{equation}
\begin{aligned}
r_i(\boldsymbol{v})
&=S_i(\widetilde{y}_i;\boldsymbol{v})-S_i(y_i;\boldsymbol{v}),\\
\Delta_i^{\mathrm{donor}}
&=r_i(\boldsymbol{v}_i^{\mathrm{donor}})-r_i(\boldsymbol{v}_i).
\end{aligned}
\label{eq:donor_shift}
\end{equation}
A negative value means donor RGB moves the donor class upward relative to the true class. We additionally report the donor's rank improvement and the fraction of predictions that turn to its label. Unlike absolute energy, none of these candidate-relative quantities can be changed by adding the same scalar offset to every candidate.

At the default timestep, matched, same-action, and different-action RGB yield $66.39\%$, $66.06\%$, and $53.12\%$, respectively. Under different-action replacement, $\Delta_i^{\mathrm{donor}}<0$ for $77.98\%$ of samples, the donor improves by 1.218 rank positions on average, and $7.80\%$ of predictions turn to the donor label. Same-action replacement changes accuracy by only $-0.34$ points, supporting semantic compatibility rather than an unconditional requirement for exact instance identity.

\paragraph{Cross-split extension.} We repeat the intervention on the three remaining evaluated protocols using the corresponding fixed main-table checkpoint for each split, $t=25$, and one shared ten-draw Gaussian noise bank across RGB conditions. Table~\ref{tab:counterfactual_crosssplit} reports 10,000-replicate class-stratified paired bootstrap intervals. Sharing the complete noise bank removes evaluation-noise variation from each paired contrast, isolating the effect of RGB substitution.

\begin{table*}[!t]
\centering
\small
\setlength{\tabcolsep}{3.5pt}
\begin{tabular}{lrrrrrrr}
\toprule
Protocol & Matched & Same & Different & M--D (95\% CI) & Donor shift (95\% CI) & Norm.\ rank gain (95\% CI) & Turn \\
\midrule
NTU-60 55/5 & 87.17 & 86.58 & 84.51 & $+2.65\ [1.70,3.69]$ & $-0.249\ [-0.262,-0.237]$ & $+0.0647\ [0.0553,0.0743]$ & 0.96 \\
NTU-120 96/24 & 70.77 & 71.02 & 64.19 & $+6.58\ [5.89,7.23]$ & $-0.329\ [-0.336,-0.323]$ & $+0.1118\ [0.1097,0.1138]$ & 5.65 \\
NTU-120 110/10 & 86.58 & 86.73 & 74.27 & $+12.31\ [11.37,13.25]$ & $-0.404\ [-0.413,-0.396]$ & $+0.1089\ [0.1061,0.1117]$ & 10.51 \\
\bottomrule
\end{tabular}
\normalsize
\caption{Cross-split controlled RGB substitutions at $t=25$. M--D is matched minus different-action top-1 in percentage points; donor shift follows Equation~\ref{eq:donor_shift}; rank gain is normalized by $K-1$; and Turn is the percentage of predictions that change to the donor label. Brackets are 95\% class-stratified bootstrap intervals.}
\label{tab:counterfactual_crosssplit}
\end{table*}

TDSM-MM consistently distinguishes matched visual evidence from action-incompatible substitutes: matched accuracy exceeds different-action accuracy on all three splits (exact two-sided McNemar $p\leq2.10\times10^{-7}$). The donor-pair shift is strictly negative and the normalized donor-rank gain is strictly positive on every split, with all corresponding confidence intervals excluding zero. Moreover, $\Delta_i^{\mathrm{donor}}<0$ for 79.65\%, 78.97\%, and 79.23\% of samples on 55/5, 96/24, and 110/10, respectively. Thus, the candidate-relative effect observed on 48/12 extends consistently to the other evaluated protocols.

\paragraph{Robustness to within-action variation.} Same-action RGB preserves performance across all three splits: the matched-minus-same differences are $+0.59$ points (95\% CI $[0.00,1.18]$, $p=0.0768$), $-0.25$ points (95\% CI $[-0.73,0.23]$, $p=0.3244$), and $-0.15$ points (95\% CI $[-0.76,0.46]$, $p=0.6909$) on 55/5, 96/24, and 110/10. Because this donor preserves the action, setup, camera, and performer while changing only the repetition, the result demonstrates robustness to within-action repetition together with the sensitivity to action semantics established by the different-action intervention. This oracle donor construction isolates semantic compatibility; during zero-shot inference, current-video RGB supplies the corresponding compatible visual observation without access to the target label.

\subsection{Dependence on Skeleton Corruption}

Table~\ref{tab:timestep_sweep} applies the same intervention over the 50-step diffusion schedule. The same skeleton, candidate order, RGB donor map, and ten noise draws are reused at every timestep. We define action-incompatible RGB damage as $D(t)=\operatorname{Acc}_{\mathrm{matched}}(t)-\operatorname{Acc}_{\mathrm{different}}(t)$. Because raw MSE scales can change with timestep, the primary cross-timestep evidence is accuracy, candidate rank, the sign of $\Delta_i^{\mathrm{donor}}$, and label transitions.

\begin{table*}[!t]
\centering
\small
\begin{tabular}{rrrrrrrrr}
\toprule
$t$ & Matched & Same & Different & $D(t)$ & $\overline{\Delta}^{\mathrm{donor}}$ & $P_{-}$ & Rank gain & Turn-to-donor \\
\midrule
1  & 56.36 & 56.79 & 54.16 & $2.20$  & $-0.113$ & 68.90 & 0.445 & 3.15 \\
5  & 59.63 & 59.54 & 55.69 & $3.94$  & $-0.157$ & 70.03 & 0.627 & 3.73 \\
10 & 62.14 & 62.39 & 55.84 & $6.30$  & $-0.202$ & 71.28 & 0.801 & 4.59 \\
20 & 66.09 & 64.98 & 54.13 & $11.96$ & $-0.307$ & 75.38 & 1.094 & 6.54 \\
25 & 66.39 & 66.06 & 53.12 & $13.27$ & $-0.363$ & 77.98 & 1.218 & 7.80 \\
35 & 65.93 & 65.32 & 48.23 & $17.71$ & $-0.477$ & 82.45 & 1.506 & 10.64 \\
49 & 62.60 & 62.20 & 39.08 & $23.52$ & $-0.588$ & 87.03 & 1.892 & 14.28 \\
\bottomrule
\end{tabular}
\normalsize
\caption{Counterfactual RGB effects across diffusion timesteps on NTU-60 48/12. $D(t)$ is matched minus different-action accuracy; $\overline{\Delta}^{\mathrm{donor}}$ is the mean donor-pair shift; $P_{-}$ is the fraction with negative shift.}
\label{tab:timestep_sweep}
\end{table*}

Different-action damage grows from 2.20 points at $t=1$ to 23.52 points at $t=49$, an endpoint increase of 21.31 points. The normalized-t slope of $D(t)$ is 21.48 points (95\% CI $[19.45,23.52]$), whereas the same-minus-matched slope is $-0.82$ points (95\% CI $[-2.29,0.66]$). Donor shifts become more frequently negative, donors move farther upward, and turn-to-donor transitions become more common. These output-level interventions support the visual-anchor interpretation: incompatible visual evidence becomes more consequential as the diffused skeleton is less informative, while semantically compatible RGB remains comparatively stable.

\paragraph{Cross-protocol endpoint extension.} We test whether this corruption-dependent behavior extends beyond 48/12 by evaluating $t\in\{1,25,49\}$ on the other three evaluated protocols. Within each protocol, the checkpoint, skeleton features, candidate texts and order, RGB donor map, and ten-noise bank are fixed across timesteps. Using the damage $D(t)$ defined above, we measure endpoint growth as $G=D(49)-D(1)$.

\begin{table*}[!t]
\centering
\small
\setlength{\tabcolsep}{7pt}
\begin{tabular}{lrrrrr}
\toprule
Protocol & $D(1)$ & $D(25)$ & $D(49)$ & $G$ & 95\% CI for $G$ \\
\midrule
NTU-60 48/12   & 2.20 & 13.27 & 23.52 & \textbf{21.31} & $[19.27,23.33]$ \\
NTU-60 55/5    & 3.91 & 2.65  & 9.73  & \textbf{5.83}  & $[3.83,7.74]$ \\
NTU-120 96/24  & 3.04 & 6.58  & 13.63 & \textbf{10.59} & $[9.74,11.43]$ \\
NTU-120 110/10 & 8.61 & 12.31 & 21.73 & \textbf{13.12} & $[11.85,14.41]$ \\
\bottomrule
\end{tabular}
\normalsize
\caption{Cross-protocol growth of action-incompatible RGB damage. $D(t)$ is matched minus different-action accuracy, and $G=D(49)-D(1)$; all values are percentage points. Intervals use 10,000 paired bootstrap replicates: class-stratified for the three endpoint extensions and the existing sample bootstrap for 48/12.}
\label{tab:timestep_crosssplit}
\end{table*}

Table~\ref{tab:timestep_crosssplit} shows positive growth on all four protocols, with every paired interval strictly above zero. On the three newly extended splits, the endpoint change in matched-minus-same-action accuracy remains within 0.38 points in magnitude, whereas incompatible-RGB damage grows by 5.83--13.12 points. From $t=1$ to $t=49$, normalized donor-rank movement and the turn-to-donor rate also increase on every split, by 6.57--19.91 and 6.78--9.25 percentage points, respectively. The endpoint effect therefore transfers across the evaluated protocols: under high rather than low skeleton corruption, the non-diffused RGB condition exerts greater candidate-relative influence, while action-compatible RGB remains stable.

\subsection{Candidate-Level Score Trajectories}

\begin{figure*}[p]
\centering
\includegraphics[width=\textwidth]{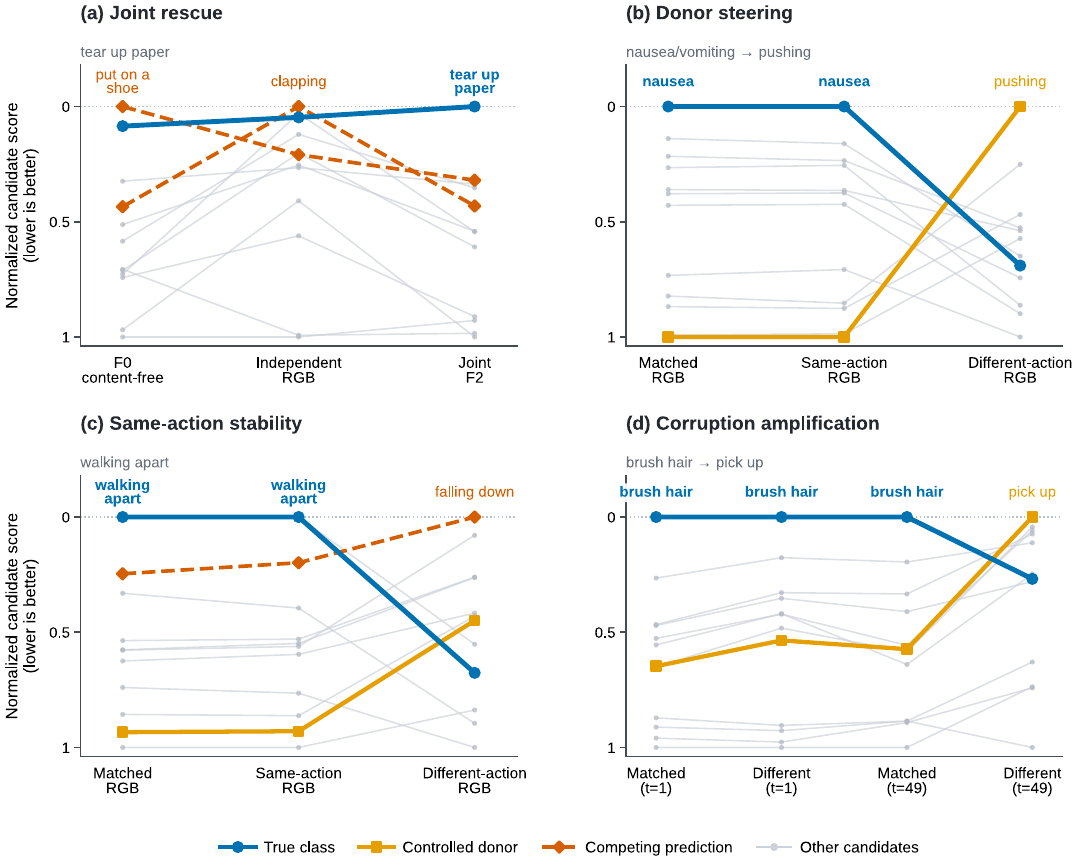}
\caption{Candidate-level score trajectories under controlled visual evidence. Scores are oriented lower-is-better and min--max normalized within each condition, so zero marks the prediction while all ranks are preserved. (a) Joint F2 recovers the true class missed by both tested independent scorers. (b) Different-action RGB promotes its donor. (c) Same-action RGB preserves the correct ordering, whereas incompatible RGB changes the winner. (d) The same donor preserves the true winner at $t=1$ but becomes the prediction at $t=49$, illustrating the aggregate corruption-dependent effect.}
\label{fig:candidate_energy_cases}
\end{figure*}

Figure~\ref{fig:candidate_energy_cases} visualizes the candidate-level mechanism underlying the aggregate results above. We reuse the per-sample score arrays from the preceding fixed-model evaluations and shared noise banks, without retraining or changing the inference protocol. Scores are oriented so that lower is better and min--max normalized across candidates within each condition; this transformation preserves the complete ranking and its top-1 prediction. High-effect cases are selected deterministically within the outcome pattern illustrated by each panel, making the candidate reordering visible while the preceding aggregate analyses quantify its prevalence.

\FloatBarrier

Panel (a) makes the correctness-overlap result concrete: joint conditioning produces the correct ordering even though neither independent branch yields the true top-1 class. Panels (b) and (c) show that changing only RGB can alter relative candidate energies according to action compatibility, rather than adding a candidate-independent offset. Panel (d) further shows that the same action-incompatible donor produces stronger reordering when skeleton corruption increases. Together with the aggregate paired results in Tables~\ref{tab:counterfactual_crosssplit} and~\ref{tab:timestep_crosssplit}, these trajectories establish that RGB participates inside candidate-wise scoring, provides action-relevant evidence, and becomes more influential as skeleton evidence is corrupted.

\subsection{Clean/Noisy Training--Testing Factorial}

The route ablation changes how RGB is presented during training, whereas the counterfactual study intervenes at inference. To separate these factors, we evaluate clean-trained and noisy-trained condition models with either a non-diffused RGB condition token or a controlled noisy RGB condition token after projection. Table~\ref{tab:factorial} reports both top-1 accuracy and the mean margin from Equation~\ref{eq:classification_margin}.

\begin{table}[t]
\centering
\small
\begin{tabular}{llrr}
\toprule
RGB training & RGB test & Top-1 & Margin \\
\midrule
Clean-trained & Clean & 66.39 & 0.1535 \\
Clean-trained & Noisy & 64.92 & 0.1189 \\
Noisy-trained & Clean & 55.32 & 0.0291 \\
Noisy-trained & Noisy & 55.23 & 0.0288 \\
\bottomrule
\end{tabular}
\normalsize
\caption{Clean/noisy RGB training--testing factorial on NTU-60 48/12.}
\label{tab:factorial}
\end{table}

The top-1 train-by-test interaction is $+1.38$ points with a 95\% CI of $[-0.06,2.87]$, so its interval narrowly includes zero. The candidate-margin interaction is $+0.0343$ with a 95\% CI of $[0.0238,0.0446]$. Clean-condition training therefore yields a clear improvement in candidate separation, while the top-1 interaction remains less conclusive.

\subsection{Class-Level Effects on NTU-120 110/10}

Table~\ref{tab:perclass_110_10} holds the 110/10 checkpoint, skeleton, candidate texts, timestep, and ten noise draws fixed while replacing RGB with a zero vector, the seen-training mean, or a real label-mismatched feature. The learned RGB projection bias and slot embedding remain present under the zero condition, so zero and mean are content-removal controls rather than retrained architectures.

\begin{table}[t]
\centering
\small
\begin{tabular}{@{}p{0.43\columnwidth}r@{\ }l@{}}
\toprule
Unseen class ($n$) & Matched & Zero / mean / mismatch \\
\midrule
drop (275) & 49.5 & 46.5 / 53.5 / 0.4 \\
put on jacket (275) & 94.9 & 25.1 / 93.1 / 11.3 \\
salute (275) & 49.5 & 87.6 / 81.1 / 86.2 \\
headache (275) & 94.2 & 9.1 / 75.6 / 11.3 \\
punch/slap (274) & 95.3 & 94.9 / 95.3 / 1.5 \\
juggling table tennis balls (514) & 86.4 & 46.7 / 0.0 / 0.0 \\
put something into bag (515) & 90.3 & 67.0 / 2.1 / 47.6 \\
cross arms (515) & 85.4 & 6.8 / 4.1 / 0.6 \\
butt kicks (514) & 97.9 & 97.5 / 91.6 / 80.9 \\
wield knife toward another person (516) & 99.4 & 45.0 / 0.4 / 83.9 \\
\midrule
\textbf{Overall (3,948)} & \textbf{86.58} & 52.58 / 40.53 / 35.49 \\
\bottomrule
\end{tabular}
\normalsize
\caption{Per-class top-1 under paired RGB-content interventions on NTU-120 110/10. Controls list zero, train-mean, and mismatched RGB, respectively.}
\label{tab:perclass_110_10}
\end{table}

Matched RGB gives a positive point estimate against all three controls for seven of ten classes; five remain supported by both subject-cluster bootstrap intervals and Holm-adjusted exact McNemar tests. The largest effects occur for classes such as \emph{headache}, \emph{juggling table tennis balls}, and \emph{cross arms}, whose skeleton patterns can be insufficiently specific. The stable regression for \emph{salute} shows that RGB can also redistribute decisions unfavorably within a visually confusable class set. The aggregate effect is therefore driven by class-dependent complementarity rather than a uniform improvement for every action.

\section{Inference Sensitivity and Computational Cost}

\subsection{Timestep and Noise-Count Sensitivity}

With matched RGB and ten noise draws, the fixed 48/12 checkpoint reaches $66.09\%$, $66.39\%$, and $65.93\%$ at $t=20$, 25, and 35, respectively. Thus, the inherited $t=25$ setting lies in a stable moderate-noise region; the sweep is a sensitivity analysis rather than a procedure for selecting $t$ from unseen accuracy.

Table~\ref{tab:noise_sensitivity} uses a fresh 20-draw bank and evaluates subsets of size $N\in\{1,5,10,20\}$. For $N=5$ and 10, we sample 1,000 subsets from the same 20 draws; $N=1$ uses all 20 singleton subsets. Monte Carlo variation falls sharply through ten draws. The nested $N=10$ and $N=20$ predictions reach $66.606\%$ and $66.575\%$, respectively, so doubling the score-generation work does not improve accuracy in this comparison.

\begin{table}[t]
\centering
\small
\begin{tabular}{rrrrr}
\toprule
$N$ & Subsets & Mean $\pm$ SD & Min--max & Cost \\
\midrule
1 & 20 & $65.28\pm1.79$ & 61.93--69.63 & 5\% \\
5 & 1,000 & $66.56\pm0.72$ & 64.47--68.59 & 25\% \\
10 & 1,000 & $66.64\pm0.44$ & 65.29--68.07 & 50\% \\
20 & 1 & 66.58 & 66.58 & 100\% \\
\bottomrule
\end{tabular}
\normalsize
\caption{Noise-count sensitivity on NTU-60 48/12.}
\label{tab:noise_sensitivity}
\end{table}

\subsection{Resource Profile}

Table~\ref{tab:resource_profile} separates the candidate scorer from offline visual encoding. Adding the RGB projection and one RGB condition token increases scorer parameters by 0.30\% and counted FLOPs by 2.69\%. At the largest profiled scorer batch, latency and allocated memory increase by 4.42\% and 4.45\%, respectively. Projecting the standard 48/12 evaluation workload of 3,270 samples, 12 candidates, and ten noise draws gives a scorer-time increase of 4.37\%.

\begin{table}[t]
\centering
\small
\begin{tabular}{@{}lrl@{}}
\toprule
Metric & TDSM $\rightarrow$ MM & Change \\
\midrule
Parameters (M) & $261.209\rightarrow261.997$ & $+0.30\%$ \\
FLOPs/candidate (G) & $3.043\rightarrow3.125$ & $+2.69\%$ \\
Batch latency (ms) & $157.14\rightarrow164.09$ & $+4.42\%$ \\
Batch memory (GiB) & $1.776\rightarrow1.855$ & $+4.45\%$ \\
Train latency (ms) & $416.54\rightarrow429.74$ & $+3.17\%$ \\
Peak memory (GiB) & $17.137\rightarrow17.634$ & $+2.90\%$ \\
48/12 scoring (s) & $120.29\rightarrow125.54$ & $+4.37\%$ \\
\bottomrule
\end{tabular}
\normalsize
\caption{Candidate-scorer resource profile. Offline feature encoders and data loading are excluded.}
\label{tab:resource_profile}
\end{table}

Precomputed RGB features occupy 4,096 bytes per video; the train and evaluation feature files used in the profile total 0.197 GiB. Encoding eight frames with the ViT-H/14 visual graph takes 55.33 ms/video, yielding 93.72 ms/video for the measured online GPU components when combined with the TDSM-MM scorer; video decoding and CPU preprocessing are excluded. This architecture-only timing uses the same ViT-H/14 computation graph and tensor shapes, although pretrained weight values were not loaded for the timing run. Consequently, the cached-feature figures quantify the architectural scorer overhead, while online deployment additionally incurs the frozen visual encoder.

\section{Summary of Supplementary Evidence}

The expanded controls show that the observed gain is not explained by an occupied token slot, simple skeleton-side RGB routing, or the tested fixed and learned independent-score fusion schemes. Across all four evaluated NTU protocols, fixed-checkpoint interventions show that RGB changes relative candidate energies according to action compatibility, and the cross-protocol endpoint extension shows that incompatible RGB is more disruptive under high than low skeleton corruption. Together with the sensitivity and resource analyses, these results characterize both the operating behavior and the practical cost of non-diffused RGB conditioning.